%% file: acl_latex.tex
\pdfoutput=1

\documentclass[11pt]{article}

\usepackage[final]{acl}

\usepackage{times}
\usepackage{latexsym}

\usepackage[T1]{fontenc}

\usepackage[utf8]{inputenc}

\usepackage{microtype}

\usepackage{inconsolata}

\usepackage{graphicx}

\usepackage{multirow}
\usepackage{booktabs}
\usepackage{amssymb}
\usepackage{colortbl}
\usepackage{makecell} 
\usepackage{multicol}
\usepackage{amsmath,amsfonts,bm}

\usepackage{float}
\usepackage[most]{tcolorbox}
\usepackage[para]{threeparttable}
\usepackage{graphicx}
\usepackage{amsmath}
\usepackage{amssymb}
\usepackage{hyperref}
\usepackage{mathtools}
\usepackage[ruled,vlined]{algorithm2e}
\usepackage{textcomp}
\usepackage{siunitx}
\usepackage{pgf}
\usepackage{subcaption}
\usepackage{xfp}

\newcommand{\OURS}{Med-PRM\xspace}
\newcommand{\RAGLLM}{\textsc{{RAG-as-a-Judge}\xspace}}
\newcommand{\GAIN}{13.50\%}
\newcommand{\RECORD}{80.35\%}

\title{\OURS{}: Medical Reasoning Models with \\Stepwise, Guideline-verified Process Rewards}
\author{
\textbf{Jaehoon Yun}$^{1,4,5*}$,  
\textbf{Jiwoong Sohn}$^{1,2}$\thanks{\quad Equal contribution.},
\textbf{Jungwoo Park}$^{1,4*}$,
\textbf{Hyunjae Kim}$^{3}$, \\
\textbf{Xiangru Tang}$^{3}$, 
\textbf{Yanjun Shao}$^{3}$,  
\textbf{Yonghoe Koo}$^{6}$, 
\textbf{Minhyeok Ko}$^{5}$, \\
\textbf{Qingyu Chen}$^{3}$, 
\textbf{Mark Gerstein}$^{3}$,  
\textbf{Michael Moor}\textsuperscript{ 2\textdagger},
\textbf{Jaewoo Kang}\textsuperscript{1,4}\thanks{\quad Corresponding authors.} \\
$^1$ Korea University, 
$^2$ ETH Zürich, 
$^3$ Yale University, 
$^4$ AIGEN Sciences, \\
$^5$ Hanyang University College of Medicine, 
$^6$ University of Ulsan College of Medicine \\
}

\begin{document}
\maketitle

\input{tabs/00_abstract}

\section{Introduction}
\label{sec:introduction}
\input{tabs/01_introduction_1}

\section{Related Work}
\label{sec:related_work}
\input{tabs/02_related_work}

\section{Preliminaries}
\label{sec:preliminaries}
\input{tabs/03_preliminaries}

\section{Method}
\label{sec:method}

\input{tabs/04_method}

\section{Experimental Setup}
\label{sec:experimental_settings}
\input{tabs/05_experimental_settings}

\section{Results}
\label{sec:results}
\input{tabs/06_results}

\section{Analysis}
\label{sec:analysis}
\input{tabs/07_analysis}

\section{Conclusion}
\label{sec:conclusion}
\input{tabs/08_conclusion}

\section*{Limitations}
\label{sec:limitations}
\input{tabs/99_limitations}

\section*{Acknowledgments}
\label{sec:acknowledgments}
\input{tabs/99_acknowledgments}

\bibliography{references_cr}

\clearpage

\appendix


\input{tabs/08_appendix}

\end{document}

%% file: tabs/00_abstract.tex
\begin{abstract}
Large language models have shown promise in clinical decision making, but current approaches struggle to localize and correct errors at specific steps of the reasoning process. This limitation is critical in medicine, where identifying and addressing reasoning errors is essential for accurate diagnosis and effective patient care. We introduce \OURS{}, a process reward modeling framework that leverages retrieval-augmented generation to verify each reasoning step against established medical knowledge bases. By verifying intermediate reasoning steps with evidence retrieved from clinical guidelines and literature, our model can precisely assess the reasoning quality in a fine-grained manner. Evaluations on five medical QA benchmarks and two open-ended diagnostic tasks demonstrate that \OURS{} achieves state-of-the-art performance, with improving the performance of base models by up to \GAIN{} using \OURS{}. Moreover, we demonstrate the generality of \OURS{} by integrating it in a plug-and-play fashion with strong policy models such as Meerkat, achieving over 80\% accuracy on MedQA for the first time using small-scale models of 8 billion parameters. Our code and data are available at \href{https://med-prm.github.io/}{Med-PRM.github.io}.
\end{abstract}

%% file: tabs/01_introduction_1.tex
Clinical decision making (CDM) is a complex, multi-step process involving the assessment of patient symptoms, retrieval of relevant clinical evidence, and formulation of diagnostic and treatment strategies. Unlike simple factual recall, CDM requires integrating diverse clinical findings and dynamically refining hypotheses as new information becomes available. Effective CDM entails not only selecting the most probable differential diagnoses but also determining what additional information is needed to reduce uncertainty and guide the next best diagnostic and therapeutic steps in a patient's clinical trajectory~\cite{moor_foundation_2023}.

While CDM spans a broad sequence of clinical decisions, a core subcomponent is the step-by-step reasoning that underlies each decision point. This sequential structure makes CDM reasoning well-suited to process reward modeling (PRM)~\cite{lightman_lets_2023, uesato_solving_2022, setlur_rewarding_2024}, which evaluates and rewards intermediate steps of a process rather than solely its final outcome. In medical practice, sound intermediate reasoning is critical to ensuring safety, reliability, and adherence to the standard of care. This creates a strong clinical motivation for models that support stepwise verification and feedback.

Recent advances in large language models (LLMs) have substantially improved performance in medical applications through pre-training~\cite{chen_meditron-70b_2023, moor_med-flamingo_2023}, post-training~\cite{kim_small_2025}, retrieval~\cite{zakka2024almanac, jeong_improving_2024, sohn_rationale-guided_2025}, tool augmentation, and agentic systems~\cite{tang_medagents_2024, schmidgall_agentclinic_2024}. More recently, emerging reasoning models~\cite{openai_introducing_2024, deepseek-ai_deepseek-r1_2025} have demonstrated the ability to decompose complex tasks into interpretable steps and exhibit meta-cognitive skills such as planning and error correction. However, such abilities are underexplored in clinical domains, where transparency, robustness, and alignment with medical standards are critical for delivering high-quality care.

Despite PRM’s potential for medical reasoning, its application to the medical domain poses key challenges. Chief among these is the need for high-quality, step-level supervision, which is both expensive and labor-intensive to obtain. Early studies~\cite{lightman_lets_2023} relied on human annotation, which is not scalable~\cite{setlur_rewarding_2024}. More recent works employed automatic labeling strategies such as Monte Carlo Tree Search (MCTS)~\cite{math_shepherd_wang}, which estimate the quality of reasoning step quality based on the probability of reaching the correct final answer from that step. Notably, MedS$^3$~\cite{jiang_meds3_2025}, a domain-specific PRM for clinical QA, also adopts an MCTS-based approach. However, these strategies often undervalue early reasoning steps that are logically sound but fail to lead to the correct outcome. This limitation is especially problematic as penalizing valid early steps can distort the learning signal and ultimately hinder the model’s ability to evaluate intermediate reasoning accurately.

Second, medical reasoning requires extensive domain knowledge that may not be fully captured within language model’s parameters alone. Thus, it necessitates a robust method to incorporate medical knowledge to generate factual, evidence-based outcomes and prevent hallucinations. In particular, training reward models solely on labels without medical context is insufficient for learning the rationale behind those labels. To overcome this, it is essential to provide relevant medical information, such as clinical guidelines, during training, enabling a more accurate interpretation of stepwise reward signals grounded in medical reasoning.

To address these challenges, we propose \OURS{}, a retrieval-augmented process reward modeling framework for clinical reasoning. Our method employs a \RAGLLM{} approach to perform stepwise evaluation conditioned on both the clinical question and retrieved medical documents. This retrieval-augmented evaluation aligns more closely with expert physician annotations than sampling-based auto-labeling methods used during training. By incorporating relevant clinical knowledge at both the training and inference stages, \OURS{} enables more accurate assessment of intermediate reasoning and outperforms existing PRM baselines by an average of 3.44\% across seven medical benchmarks.

Our experiments demonstrate that test-time scaling, where \OURS{} is used as a verifier alongside a fine-tuned policy model, achieves state-of-the-art performance. Moreover, \OURS{} exhibits strong plug-and-play generality, achieving further gains when applied to top-performing models such as UltraMedical~\cite{zhang_ultramedical_2024}. In contrast to UltraMedical, which was trained on data costing approximately \$20,000, our reward model relies on a curated dataset costing less than \$20, highlighting both the cost-efficiency and scalability of our approach.

Our contributions are as follows: 
\begin{enumerate}
\item We propose \OURS{}, a retrieval-augmented process reward modeling framework that evaluates each reasoning step in the context of both the clinical question and retrieved evidence, enabling fine-grained and evidence-grounded assessment. 
\item We demonstrate that \OURS{} achieves state-of-the-art performance across six out of seven medical QA benchmarks, outperforming all baseline language, reasoning, and medical models. Our verifier improves base model performance by up to \GAIN{} at test time and reaches \RECORD{} on MedQA (4 options) using only 8B-parameter models.
\item Through in-depth qualitative analysis and collaboration with medical experts, we show that \OURS{} closely aligns with clinical experts, addressing key limitations of prior training methods of PRMs in both logical consistency and factual accuracy.
\end{enumerate}

%% file: tabs/02_related_work.tex
For a more detailed overview of related work, refer to Appendix~\ref{supp:related}. LLMs have shown increasing proficiency in medical reasoning, effectively handling domain-specific terminology and multimodal data. Using corpora like PubMed, MIMIC-III/IV~\cite{johnson_mimic-iv_2023}, and UMLS, recent models go beyond surface-level recall to complex inference. 

Med-PaLM~\cite{singhal_large_2023} demonstrates promising performance on expert-level medical QA benchmarks, while methods such as CoT prompting~\cite{wei_chain--thought_2023, xu_mental-llm_2024, kim_small_2025}, agentic frameworks~\cite{kim_mdagents_2024, tang_medagents_2024, schmidgall_agentclinic_2024}, and PRM~\cite{jiang_meds3_2025} further enhance reasoning. Reinforcement learning~\cite{math_shepherd_wang} and verifier feedback~\cite{chen_huatuogpt-o1_2024} have been applied to refine reasoning traces, emphasizing the need for stepwise supervision. 

MedS$^3$\cite{jiang_meds3_2025} applies PRM using MCTS-based auto-labeling. Our approach, \OURS{}, also leverages process-level rewards but differs by incorporating retrieval-augmented generation and an LLM-as-a-Judge framework. As shown in Section~\ref{sec:results} and Section~\ref{sec:analysis}, this yields superior performance compared to MCTS-based methods.

\citet{hao_llm_2024} explore the use of LLMs as verifiers for CoT reasoning. \OURS{} adopts \RAGLLM{} for stepwise supervision, diverging from earlier PRM approaches that rely on automatic scoring. In mathematics, RAG-PRM~\cite{zhu_retrieval-augmented_2025} has been proposed to retrieve similar QA pairs for few-shot prompting with PRM. In contrast, \OURS{} retrieves medical knowledge and evidence, enabling integration of diverse sources like textbooks or clinical guidelines.

%% file: tabs/03_preliminaries.tex
\input{figs/overview}
\subsection{Reward Model}
Reward models have emerged as central to advancing LLMs beyond pre-training and fine-tuning, driven by two key developments in reinforcement learning (RL) and test-time compute scaling. RL methods such as Proximal Policy Optimization (PPO) rely on carefully designed reward functions to optimize model behavior in settings where ground-truth supervision is sparse or costly. Additionally, test-time strategies like best-of-N~\cite{lightman_lets_2023} have proven highly effective as alternatives to majority voting, such as self-consistency~\cite{wang_self-consistency_2023}, using reward models to rank and select high-quality outputs. These trends clearly highlight the growing importance of accurate, context-aware reward models not only during training but also at inference time.

\paragraph{Outcome Reward Model (ORM)}
Given a question $q$ and a model-generated reasoning trace $S$ the ORM assigns a sigmoid score $r_S \in [0,1]$ indicating the correctness of the entire trace. ORM is trained with the following cross-entropy loss:
\begin{align*}
\mathcal{L}_{\text{ORM}} = - \left( y_S \log r_S + (1 - y_S) \log(1 - r_S) \right),
\end{align*}
where $y_S$ is the gold label of the reasoning trace $S$, $y_S = 1$ if $S$ is correct, and $y_S = 0$ otherwise. 

\paragraph{Process Reward Model (PRM)}
Given a reasoning trace $S = (s_1, s_2, \cdots, s_K)$ where $K$ is the number of reasoning steps, a PRM assigns score $r_{s_i} \in [0,1]$ for each step $s_i$. Gold labels $y_{s_i} \in \{0,1\}$ indicate whether each step is correct. To compute these scores, the model predicts logits for the special tokens \texttt{+} (correct) and \texttt{-} (incorrect) appended to each reasoning step. The confidence score $r_{s_i}$ is defined as the softmax probability of the \texttt{+} token over the logits of both tokens. The model is trained to minimize the cross-entropy loss over all reasoning steps:
\begin{align*}
\resizebox{0.95\columnwidth}{!}{$
\mathcal{L}_{\text{PRM}} = -\sum_{i=1}^{K} \left( y_{s_i} \log r_{s_i} + (1 - y_{s_i}) \log(1 - r_{s_i}) \right)
$}
\end{align*}

PRM takes as input the concatenation of the question \(q\) and the reasoning trace \(S\), and produces stepwise confidence scores as follows: \((r_{s_1},\ r_{s_2},\ \cdots,\ r_{s_K})\) And the minimum step score is defined as the score of the solution S:
\begin{align*}
\resizebox{0.95\columnwidth}{!}{$
\text{RM}(q, S) = r_S, \ \text{where } r_S = \min(r_{s_1},\ r_{s_2},\ \cdots,\ r_{s_K})
$}
\end{align*}

\paragraph{PRM Auto-Labeling}
\citet{math_shepherd_wang} proposed an auto-labeling method to address the cost of human annotations $y_{s_i}$. Inspired by MCTS, a completer model generates N subsequent reasoning processes from each partial trace up to step $s_i$, producing sequences of the form$
\left\{ (s_{i+1,j}, \cdots, s_{{K_j},j}, a_j) \right\}_{j=1}^{N}$,
where $a_j$ is the final answer of the $j$-th continuation. Let $a^*$ denote the gold answer to the question $q$. A hard label is assigned to step $s_i$ as follows:
\begin{align*}
y^{HE}_{s_i} =
\begin{cases}
1 & \text{if } \exists j \ \textrm{such that} \ a_j = a^*, \\
0 & \text{otherwise.}
\end{cases}
\end{align*}

A soft label is computed as the empirical probability of reaching the correct answer:
\begin{align*}
y^{SE}_{s_i} = \frac{1}{N} \sum_{j=1}^{N} \mathbb{I}(a_j = a^*)
\end{align*}

This auto-labeling approach enables scalable PRM training without human supervision. However, it is prone to false negatives---particularly in complex questions---where factually correct and logically coherent intermediate steps are mislabeled as incorrect (or assigned low soft-label scores) simply because none of their sampled continuations lead to the correct final answer. To mitigate false negatives, we incorporate retrieval-based fact-checking into the labeling process, as described in Section \ref{sec:method}.

%% file: figs/overview.tex
\begin{figure*}[t]
    \centering
    \includegraphics[width=0.82\textwidth]{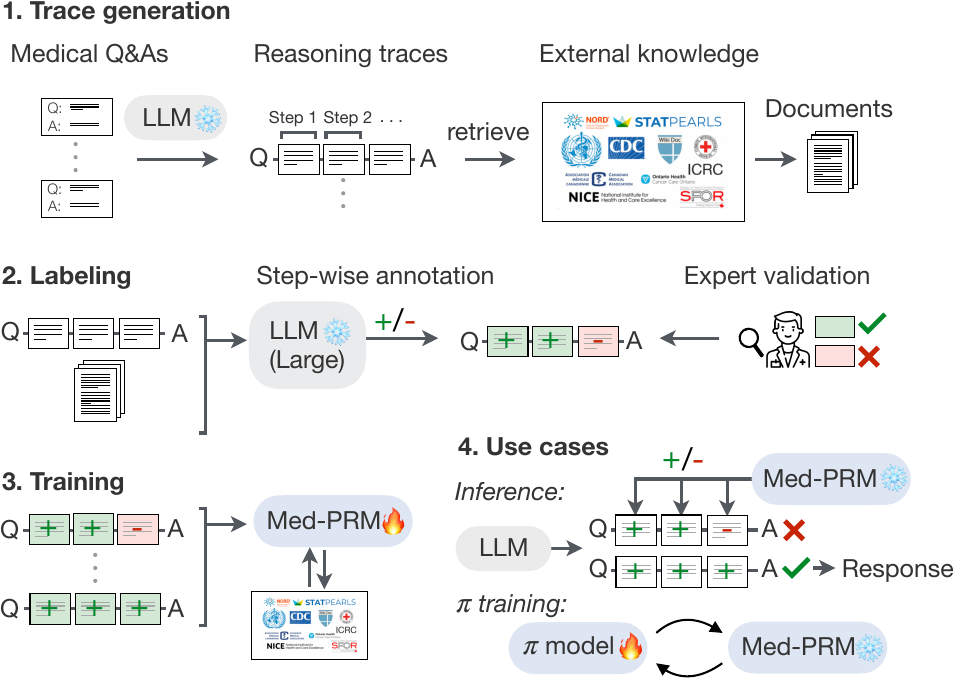} 
    \caption{    Overview of the \textsc{Med-PRM}. (1) An LLM generates reasoning traces for medical questions, and relevant documents are retrieved from external corpora.  
    (2) A large LLM assigns stepwise labels (+/–) for each reasoning step.
    (3) These labeled traces are used to train the Med-PRM reward model.  
    (4) Med-PRM reward model is used for inference-time evaluation or further train the policy model.
    }
    \label{fig:overview}
\end{figure*}

%% file: tabs/04_method.tex
\subsection{\RAGLLM{}}

\label{subsec:ours-training}We use \RAGLLM{} with a labeling strategy that differs from conventional PRMs. Given a question $q$, golden answer $a^*$, a set of retrieved documents $D$, and a sequence of reasoning steps $S = (s_1, \ldots, s_K)$, \RAGLLM{} performs binary classification on each step $s_i$ to determine whether it is correct, producing labels $y^{RAG}_{s_i} \in \{0, 1\}$:
\[
\resizebox{0.95\columnwidth}{!}{$
\RAGLLM(D,q,a^*,S)= y^{RAG}_{S}$
}
\]
These labels $y^{RAG}_{S}$ are plugged in $\mathcal{L}_{\textrm{PRM}}$ in order to train our PRM. Each score $r_{s_i}$ reflects the likelihood that step $s_i$ is correct, following~\citet{lightman_lets_2023}.
During both training and evaluation, PRM receives the same input: question $q$, retrieved documents $D$, and reasoning trace $S$. The key difference from prior PRM models is the inclusion of the retrieved documents \(D\) as part of the input.
\begin{align*}
\text{RM}(D, q, S) &= r_S
\end{align*}

To construct the retrieval query, we concatenate the question with the reasoning trace:
\[
    D = \text{Retriever}(q,S)
\]

During training, the correct reasoning trace among multiple inferences is used as the query. During inference, a randomly sampled reasoning trace is used as the query instead. Further implementation details of document retrieval are provided in Appendix~\ref{appendix:retrieval}.

\paragraph{Scaling Test-time Computation}
Following \citet{lightman_lets_2023}, we define a reasoning trace's final score as the minimum reward across its steps. The best-of-N approach selects an answer among the traces with the highest final score. Let $C = \{S_1, S_2, \cdots, S_N\}$ be the set of reasoning traces and $\{a_{S_1}, a_{S_2}, \cdots, a_{S_N}\}$ the corresponding answers. Then, we have
\begin{align*}
a_{rm} = a_{S_*}, \quad \text{where } S_* = \arg\max_{S_j \in C} r_{S_j},
\end{align*} and where $r_{S_j}$ is a final score of reasoning trace ${S_j}$
We also adopt a hybrid method following \citet{li_making_2023} that combines self-consistency and reward scoring, noted as SC+RM (Self-Consistency + Reward Model). Traces are grouped by their final answers, and the answer with the highest total reward is selected:
\begin{align*}
a_{sc+rm} = \arg\max_a \sum_{j=1}^N \mathbb{I}(a_{S_j} = a) \cdot \mathrm{RM}(q, S_j),
\end{align*} 
where $\mathrm{RM}(q,S_j)$ is the reward score of the $j$-th reasoning trace assigned by PRM for question $q$. 
A strong reward model improves selection by assigning higher scores to correct reasoning traces.

\subsection{Policy Model Fine-tuning}
We fine-tune the policy model using a rejection sampling guided by \OURS{}. For each question in the training set, multiple reasoning traces are generated, and \OURS{} assigns stepwise reward scores. Traces are ranked by their minimum step score, and only top-ranked traces are retained for supervised fine-tuning.

Following~\citet{qwen_qwen25_2025}, we exclude questions consistently answered correctly to concentrate training on more challenging examples. After fine-tuning, \OURS{} is again used at inference time to rescore multiple generations and select the best one. This bootstraps the policy model to produce reasoning paths aligned with \OURS{}, improving performance on complex medical QA tasks.

%% file: tabs/05_experimental_settings.tex
\input{tables/table1_single}

\subsection{Training of \OURS{}}
\paragraph{Model}
We perform full fine-tuning of the Llama-3.1-8B-Instruct model on a single NVIDIA A100 (80GB VRAM) with a maximum sequence length of 4096 tokens. We use the AdamW optimizer with a learning rate of $2 \times 10^{-6}$, cosine decay, and 5\% warmup ratio. Training is performed in bfloat16 precision with gradient checkpointing and Flash Attention V2 with gradient accumulation to global batch size of 64. We designate the EOS token as the padding token and introduce a special marker to segment reasoning steps for process-level supervision. More details on the hyperparameters used are described in Appendix~\ref{appendix:hyperparameters}.

\paragraph{Data Filtering and Labeling}
Training uses MedQA~\cite{jin_what_2020}, MedMCQA~\cite{pal_medmcqa_2022}, PubMedQA~\cite{jin_pubmedqa_2019}, and MMLU~\cite{hendrycks_measuring_2021}. We use the full MedQA training set (10,178 questions) and sample 500 instances from each of the remaining datasets. For each question, we sample 16 candidate reasoning traces and filter out traces with fewer than three or more than nine reasoning steps to avoid overly shallow or degenerate reasoning. To maintain label balance, the number of correct reasoning traces per question was limited to no more than the number of incorrect traces or two, whichever is greater. To ensure the total number of tokens to not exceed 4096, 1024 tokens were reserved for the question and reasoning, and the remaining 3072 tokens were used to sequentially include relevant documents. Retrieved documents (truncated to 3072 tokens) are prepended to the question and a reasoning trace with each step separated using a special token to form the input. stepwise binary supervision is applied at each marker using labels from \RAGLLM{}.

\subsection{Evaluation of \OURS{}}

\paragraph{Benchmarks}
We evaluate \OURS{} on MedQA (4 and 5 options), MedMCQA (validation set), six medical MMLU subsets~\cite{singhal_large_2023}, DDXPlus~\cite{tchango_ddxplus_2022}, and two open-ended AgentClinic variants~\cite{schmidgall_agentclinic_2024} based on NEJM and MedQA. AgentClinic adopts an open-ended format and is evaluated using Gemini-2.0-flash. Detailed descriptions of benchmarks are in Appendix~\ref{appendix:benchmarks}.

\paragraph{Baselines}
We benchmark \OURS{} against proprietary and open-source models, including general-purpose, reasoning, medical, and process reward models. Table~\ref{tab:table_1} summarizes performance across both multiple-choice and open-ended question answering benchmarks widely used in the medical domain. 

We further compare our method against internal baselines in the ablation study: (1) PRM\textsubscript{soft} and (2) PRM\textsubscript{hard}, both using MCTS-style auto-labeling~\cite{math_shepherd_wang}, and (3) \OURS{} without retrieval. Our final method, (4) \OURS{} with retrieval, is also included for comparison. Each baseline is evaluated under two strategies: Best-of-N and SC+RM.

\paragraph{Scaling Test-Time Computation}
We gradually increase the number of reasoning traces generated by the policy model up to $N = 64$ per question. The final answer is selected using two strategies: Best-of-N, which chooses the trace with the highest $r_{S_j}$ score, and SC+RM, which selects the answer group with the highest sum total reward score.

%% file: tables/table1_single.tex
\begin{table*}[t]
    \centering
    \setlength{\tabcolsep}{1pt} 
    \scriptsize 
    \resizebox{\textwidth}{!}{
    \begin{tabular}{ll*{9}{>{\centering\arraybackslash}p{1.4cm}}}
    \toprule
    &
    &
    & \multicolumn{5}{c}{\textbf{Multiple-Choice QA}} 
    & \multicolumn{2}{c}{\textbf{Open-Ended QA}} 
    &
    \\
    \cmidrule(lr){4-8} \cmidrule(lr){9-10}
    \makecell{\textbf{Category}} & \makecell{\textbf{Model}} & \makecell{\textbf{Size}} 
    & \textbf{MedQA-4}& \textbf{MedQA-5} & \textbf{MedMCQA} & \textbf{MMLU-Med} & \textbf{DDXPlus} 
    & \makecell{\textbf{Agent Clinic}\\\textbf{NEJM *}} & \makecell{\textbf{Agent Clinic}\\\textbf{MedQA *}} 
    & \makecell{\textbf{Average}}
    \\
    \midrule
    \multirow{3}{*}{\makecell[c]{Proprietary\\Language Models}}
        & Gemini Flash 2.0 & -- & 87.51 & 85.23 & 72.60 & 92.01 & 75.00 & 70.83 & 87.74 & 81.56 \\
        & GPT-4o-mini &--& 79.03 & 74.31 & 68.20 & 87.79 & 76.00 & 58.33 & 79.44 & 74.73 \\
        & GPT-3.5 turbo & -- & 69.91 & 65.44 & 57.00 & 76.77 & 73.80 & 51.72 & 77.93 & 67.51 \\
    \midrule
    \multirow{2}{*}{\makecell[c]{Proprietary\\Reasoning Models}}
        & o4-mini & -- & 93.95 & 91.12 & 79.60 & 93.99 & 79.80 & 76.67 & 94.86 & 87.14 \\
        & o3-mini & -- & 92.69 & 90.97 & 75.50 & 93.01 & 79.80 & 78.33 & 96.26 & 86.65 \\
    \midrule
    \midrule
    \multirow{3}{*}{\makecell[c]{Open-source\\Language Models}}
        & Llama3.1 & 8B & 70.93 & 65.20 & 61.60 & 78.97 & 68.80 & 35.83 & 71.96 & 64.76 \\
        & Gemma 2 & 9B & 64.73 & 60.25 & 53.00 & 77.87 & 64.40 & 41.67 & 66.36 & 61.18\\
        & Ministral & 8B & 56.17 & 50.43 & 49.20 & 67.22 & 51.80 & 34.17 & 62.62 & 53.09\\
    \midrule
    \multirow{7}{*}{\makecell[c]{Open-source\\Reasoning Models}}
        & DeepSeek-R1 & 671B & 90.34 & 89.87 & 78.80 & 94.40 & 79.60 & 79.83 & 91.12 & 86.28 \\
        & QwQ~\nocite{team_qwq_2024} & 32B & 85.31 & 81.62 & 70.20 & 88.89 & 74.00 & 63.33 & 85.51 & 78.41 \\
        & Sky-T1-Preview~\nocite{li_llms_2025} & 32B & 77.77 & 73.53 & 66.20 & 88.34 & 74.00 & 53.33 & 81.31 & 73.50\\
        & R1-Distill-Llama & 8B & 34.96 & 30.16 & 43.60 & 64.19 & 36.80 & 30.83 & 57.01 & 42.51 \\
        & R1-Distill-Qwen\nocite{deepseek-ai_deepseek-r1_2025} & 7B & 24.82 & 19.56 & 36.40 & 47.47 & 36.80 & 8.33 & 35.51 & 29.84\\
        & Sky-T1~\nocite{li_llms_2025} & 7B & 34.09 & 30.64 & 36.20 & 53.17 & 47.40 & 6.67 & 29.91 & 34.01\\
        & Marco-o1~\nocite{zhao_marco-o1_2024} & 7B & 39.36 & 34.80 & 49.20 & 69.15 & 38.40 & 30.83 & 63.55 & 46.47 \\
    \midrule
    \multirow{5}{*}{\makecell[c]{Open-source\\Medical Models}}
        & TX-Gemma~\nocite{wang_txgemma_2025} & 9B & 41.56 & 35.11 & 36.00 & 52.34 & 57.80 & 20.00 & 50.00 & 41.83 \\
        & Meditron3~\nocite{chen_meditron-70b_2023} & 8B & 59.94 & 52.95 & 48.20 & 67.86 & 67.80 & 42.50 & 67.29 & 58.08 \\
        & Meerkat~\nocite{kim_small_2025} & 8B & 71.25 & 69.13 & 56.40 & 76.40 & 70.00 & 43.33 & 76.40 & 66.13\\
        & UltraMedical~\nocite{zhang_ultramedical_2024} & 8B & 72.66 & 68.34 & 62.60 & 79.61 & 72.60 & 45.83 & 70.56 & 67.46\\
        & HuatuoGPT-o1~\nocite{chen_huatuogpt-o1_2024} & 8B & 72.19 & 63.24 & 63.60 & 75.30 & 64.00 & 40.00 & 71.50 & 64.26 \\
    \midrule
    \multirow{7}{*}{\makecell[c]{Open-source\\Medical Process\\Reward Models}}
        & MedS$^3$~\nocite{jiang_meds3_2025} & & & & & & & & & \\
        & \qquad Best-of-N & 8B & 71.56 & 68.42 & \underline{64.20} & 80.18 & 75.40 & 52.50 & 74.30 & 69.51 \\
        & \qquad SC+RM & 8B &
        75.64 & 71.41 & \underline{64.20} & 81.79 & 74.60 & \textbf{55.00} & 74.77 & 71.06 \\
    & \cellcolor{gray!15}{\textbf{\OURS{}}} & \cellcolor{gray!15}
        & \cellcolor{gray!15} & \cellcolor{gray!15} & \cellcolor{gray!15} & \cellcolor{gray!15} 
        & \cellcolor{gray!15} & \cellcolor{gray!15} & \cellcolor{gray!15} & \cellcolor{gray!15} \\
    & \qquad \cellcolor{gray!15}Best-of-N & \cellcolor{gray!15}8B
        & \cellcolor{gray!15}\underline{76.76} & \cellcolor{gray!15}\underline{72.43} & \cellcolor{gray!15}\underline{64.20} & \cellcolor{gray!15}\underline{82.37} 
        & \cellcolor{gray!15}\textbf{77.80} & \cellcolor{gray!15}\underline{54.17} & \cellcolor{gray!15}\textbf{80.37}& \cellcolor{gray!15}\underline{72.59} \\
    & \qquad \cellcolor{gray!15}SC+RM & \cellcolor{gray!15}8B
        & \cellcolor{gray!15}\textbf{79.18} & \cellcolor{gray!15}\textbf{75.49} & \cellcolor{gray!15}\textbf{67.40} & \cellcolor{gray!15}\textbf{83.29} 
        & \cellcolor{gray!15}\underline{77.20} & \cellcolor{gray!15}{52.50} & \cellcolor{gray!15}\underline{79.44} & \cellcolor{gray!15}\textbf{73.50} \\

    \bottomrule
    \end{tabular}
    } 
\caption{
Accuracy of proprietary and open-source models across multiple-choice and open-ended medical QA benchmarks. We use instruction-tuned models for Llama3.1, Gemma2, and Ministral. Best scores are shown in \textbf{bold}, and second-best scores are \underline{underlined} among small-scale models ($<$~10B parameters). 
We report results on AgentClinic*, simplified variants of the original benchmarks (see Appendix \ref{appendix:benchmarks} for details).
}
    \label{tab:table_1}
\end{table*}

%% file: tabs/06_results.tex
\subsection{Main Results}

Table~\ref{tab:table_1} summarizes the performance of \OURS{} compared to several baselines across seven medical benchmarks. We evaluate \OURS{} using two test-time strategies: Best-of-N and SC+RM, achieving average accuracy of 72.59\% and 73.50\%, respectively. These results outperform all existing open-source language, reasoning, medical, and medical process reward models with fewer than 10 billion parameters. With SC+RM, \OURS{} achieves state-of-the-art results on 4 out of 7 benchmarks and on 2 out of 7 benchmarks under the Best-of-N setting.

We observe larger performance improvements on benchmarks requiring complex clinical reasoning compared to knowledge-centric tasks such as MedMCQA. Notably, on the AgentClinic benchmark, which closely mirrors real-world diagnostic workflows, \OURS{} achieves accuracy gains of 12.50\% and 10.75\% under the SC+RM and Best-of-N settings, respectively.

Compared to MedS$^3$, the previous state-of-the-art process reward model at the 8B scale, \OURS{} achieves an average improvement of 2.44\% across all benchmarks using the SC+RM strategy. Even under the Best-of-N strategy, \OURS{} outperforms MedS$^3$ by 3.08\%.

These results demonstrate the strong capability of \OURS{} in identifying clinically sound reasoning paths. We further explore its effectiveness when paired with stronger, fine-tuned language models such as Meerkat-8B and UltraMedical-8B in Section~\ref{subsec:policy_reward}

\subsection{Reward Model as Verifier}
\label{subsec:policy_reward}
To assess the model-agnostic utility of \OURS{}, we apply it as a plug-and-play verifier during inference across various policy models on MedQA. As shown in Table~\ref{tab:policy_reward}, \OURS{} consistently improves performance, regardless of the underlying base or fine-tuned model.

\input{tables/policy_reward}

To further demonstrate the effectiveness of Med-\linebreak
PRM, we train a policy model using supervised\linebreak
\input{figs/medical_benchmark}
\noindent fine-tuning (SFT) on a rejection-sampled dataset constructed with our reward model, following the Entropy-Regularized PRM~\cite{zhang_entropy-regularized_2024}. 
This policy model, denoted as \OURS{} $\pi$, leverages high-quality reasoning traces selected by \OURS{} and achieves 79.18\% accuracy on MedQA with 10.39\% improvement over the base Llama-3.1-8B-Instruct model alone.

Moreover, when \OURS{} is used as a verifier on top of strong, fine-tuned models such as Meerkat-8B and UltraMedical-8B, we observe additional gains. Notably, pairing \OURS{} with Meerkat-8B yields 80.35\% accuracy on MedQA, marking the first time that an 8B-scale model has surpassed the 80\% threshold on this benchmark. These results highlight the generalizability of our reward model as a plug-and-play component for enabling more accurate medical reasoning across diverse models. A comprehensive table of full benchmark results, including CoT, SC, and PRM baselines (hard label, soft label, MedS$^3$, and \OURS{}) is provided in Appendix~\ref{appendix:Scaling Test-Time Computation with PRMs Across Multiple Models}.

%% file: tables/policy_reward.tex
\begin{table}[H]
    \centering
    \footnotesize
    \resizebox{\columnwidth}{!}{%
    \begin{tabular}{ll}
    \toprule
    \multicolumn{1}{c}{\textbf{Model}} & \multicolumn{1}{c}{\textbf{MedQA (4 options)}} \\
    \midrule
    Llama-3.1-8B-Instruct & 68.79 \\
    \quad + SC & 74.86 {(+6.07)}\\
    \rowcolor[gray]{0.9} \quad + SC + RM (\OURS{} $RM$) &\textbf{78.24} {(+9.45)} \\
    \rowcolor[gray]{0.9} \quad + Best-of-N (\OURS{} $RM$) & 76.98 {(+8.19)} \\
    \midrule
    Llama-3.1-8B-Instruct* (\OURS{} $\pi$) & 67.22 \\
    \quad + SC & 75.02{(+7.80)} \\
    \rowcolor[gray]{0.9} \quad + SC + RM (\OURS{} $RM$) & \textbf{79.18} (+11.96) \\
    \rowcolor[gray]{0.9} \quad + Best-of-N (\OURS{} $RM$) & 76.76 (+9.54) \\
    \midrule
    UltraMedical-8B$^\dagger$ & 67.51 \\
    \quad + SC & 75.63 (+8.12)\\
    \rowcolor[gray]{0.9} \quad + SC + RM (\OURS{} $RM$) & \textbf{79.87} {(+12.36)} \\
    \rowcolor[gray]{0.9} \quad + Best-of-N (\OURS{} $RM$) & 76.42 {(+8.91)} \\
    \midrule
    Meerkat-8B$^\dagger$ & 66.65 \\
    \quad + SC & 76.04 {(+9.39)}\\
    \rowcolor[gray]{0.9} \quad + SC + RM (\OURS{} $RM$) & \textbf{80.35} (+13.70) \\
    \rowcolor[gray]{0.9} \quad + Best-of-N (\OURS{} $RM$) & 79.95 (+13.30) \\
    \bottomrule
    \end{tabular}%
    }
\caption{Performance improvements from using the \OURS{} reward model as a verifier on MedQA (4 options). For each policy model, the first row shows the average score over 64 sampled solutions. Subsequent rows apply Self-Consistency (SC), SC with reward model verification (SC+RM), and Best-of-N using the same 64 solutions.}
    \label{tab:policy_reward}
    \end{table}

%% file: figs/medical_benchmark.tex
\begin{figure*}[t!]
    \centering
    \begin{subfigure}[t]{0.46\textwidth}
        \centering
        \includegraphics[width=\textwidth]{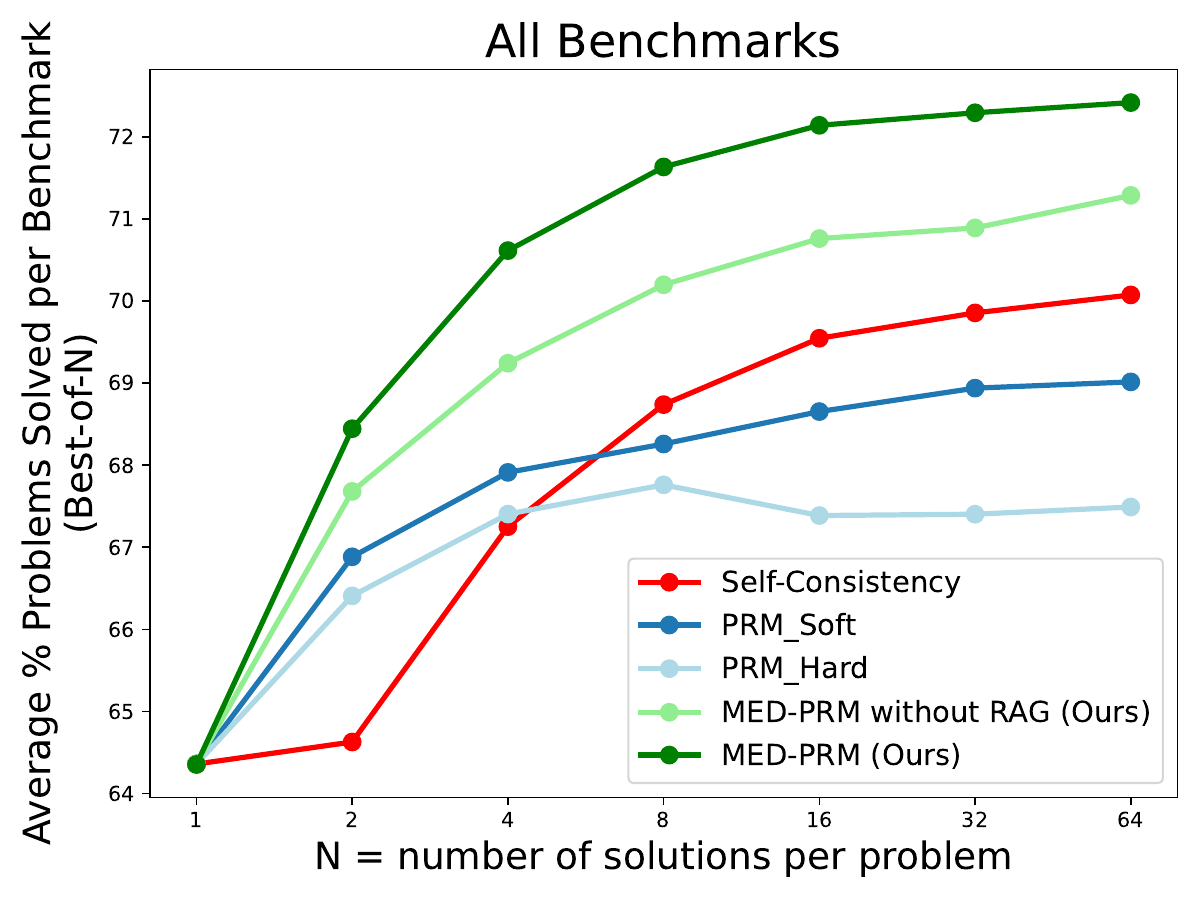}
        \caption{Best-of-N}
    \end{subfigure}
    \hfill
    \hspace{0.02\columnwidth} 
    \begin{subfigure}[t]{0.46\textwidth}
        \centering  
        \includegraphics[width=\textwidth]{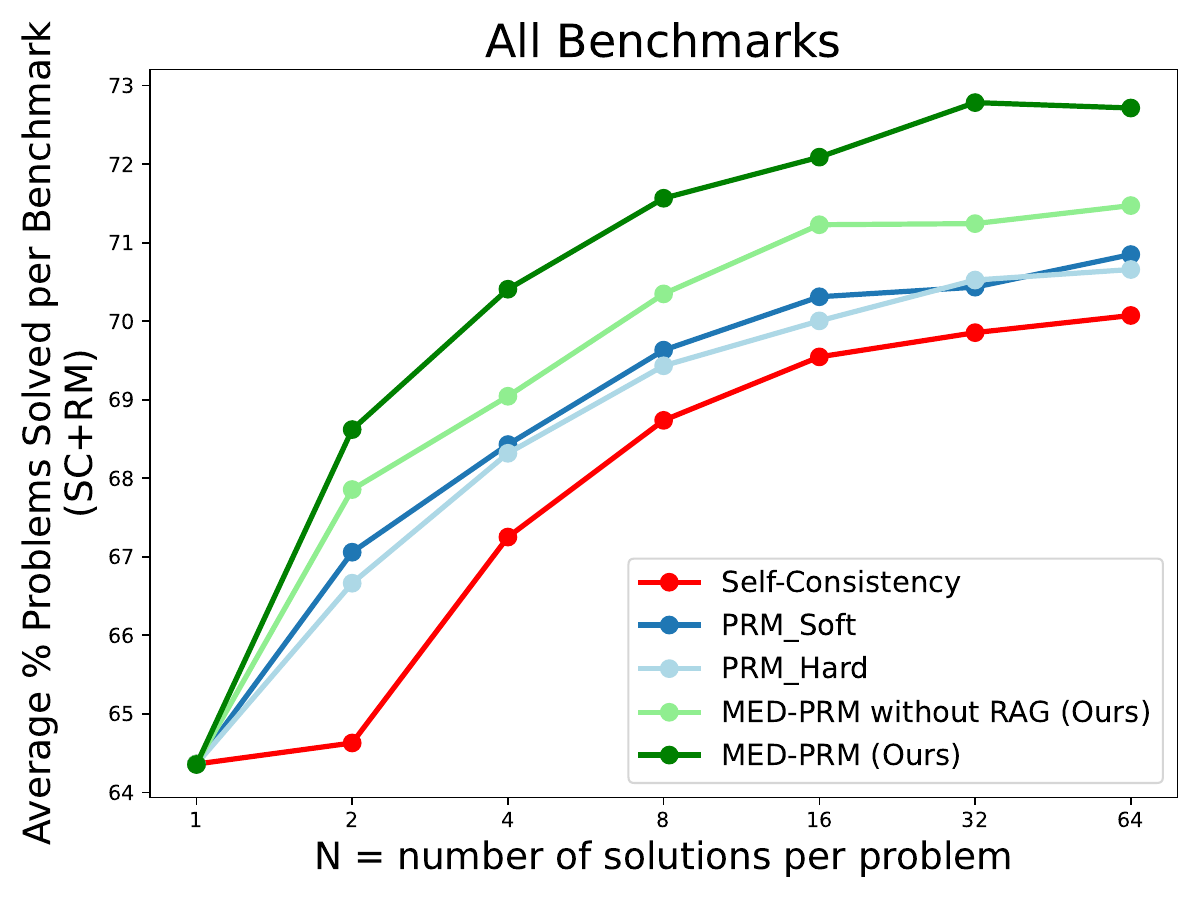}
        \caption{SC+RM}
    \end{subfigure}
    
    \caption{Comparison of scaling test-time computation performance between \OURS{} and conventionally trained PRMs across overall medical benchmarks.}
    \label{fig:ablation}
\end{figure*}

%% file: tabs/07_analysis.tex
\subsection{Ablation Study}
We conduct a comprehensive ablation study to evaluate the contribution of each component in \OURS{}. The results are visualized in Figure~\ref{fig:ablation}. We assess performance improvements under test-time scaling using 64 sampled solutions from the Llama-3.1-8B-Instruct model, evaluated with both Best-of-N and SC+RM strategies across the benchmark.

We compare \OURS{} to Self-Consistency and two PRMs trained with conventional auto-labeling methods: \textsc{PRM}\textsubscript{soft} and \textsc{PRM}\textsubscript{hard}. The results including \( \text{MedS}^{3}\) are provided in Appendix~\ref{appendix:ablation study}.

To analyze the impact of our design, we incrementally add key components. First, we replace conventional labeling with using an LLM to directly evaluate the reasoning steps. This is used to train a variant named ``\OURS{} without RAG''. Second, we incorporate retrieval for the full \OURS{} framework. 

Results show that each component yields consistent gains across both test-time strategies. Notably, \OURS{} without retrieval already outperforms conventional PRMs, and adding retrieval further boosts performance. This demonstrates the critical role of grounding in external knowledge.

\input{figs/agent_clinic} 

\input{figs/labeling_strategies_comparison}
Under the SC+RM setting, conventional PRMs achieve modest improvements over Self-Consistency. However, in the more stringent Best-of-N setting, where only the top solution is selected, conventional PRMs underperform relative to Self-Consistency. In contrast, \OURS{} consistently outperforms Self-Consistency in both settings, underscoring that LLM-based step-level supervision is more effective than sampling-based auto-labeling. These findings also suggest that Best-of-N serves as a more discriminative setting for comparing reward model performance.

\subsection{Open-Ended Clinical Tasks}
We evaluate \OURS{} on AgentClinic* with Best-of-N strategy, a diagnostic benchmark designed in an open-ended QA format to closely resemble real-world clinical settings.

Although this dataset was not introduced during the training phase, \OURS{} achieves a significant 11.81\% improvement in accuracy through scaling test-time computation. 
It further outperforms other baseline methods by a clear margin, achieving 4.87\% higher accuracy than Self-Consistency and 4.32\%
higher than PRMs trained with conventional approaches.

While multiple-choice question answering (MCQA) benchmarks provide a convenient measure of LLM performance, they do not fully capture real-world medical scenarios. Open-ended diagnostic tasks more closely reflect practical settings; however, scaling test-time strategies such as majority voting are less applicable, as no predefined answer options exist. PRMs provide a promising solution in this context by selecting the most clinically plausible answer rather than relying on option-based voting, making them particularly effective for open-ended medical reasoning.

\subsection{Expert Alignment}
\label{subsec:expert_alignment}

To assess the alignment between \OURS{} and medical experts, we calculate the Pearson correlation between model-generated labels and expert annotations on step-level reasoning quality. We select three questions each from an easy subset (where Llama-3.1-8B-Instruct achieves over 10\% accuracy) and a hard subset (accuracy below 10\%) from the PRM training set. For each question, five model-generated reasoning traces were annotated by human experts, resulting in 180 step-level annotations in total.

As shown in Table \ref{tab:human_annotation}, \OURS{} shows high correlation with human judgments across both easy and hard subsets (0.74 and 0.71, respectively). In contrast, the performance of soft and hard labeling strategies drops significantly on hard examples, dropping from 0.64 to 0.34 and 0.70 to 0.31, respectively. This suggests that \OURS{} produces more robust and consistent labels even in more challenging reasoning scenarios, making it better suited for constructing high-quality training sets.

\input{tables/human_annotation.tex}

\input{figs/comparison_scoring}

\subsection{Case Study}
\label{subsec:case_study}
\paragraph{Training Data Labeling}
Figure~\ref{fig:labeling_strategies} presents an example from the MedQA training dataset concerning a patient suspected of having Graves’ ophthalmopathy. Although diplopia and ocular pain are commonly associated with Graves’ disease, they result from autoimmune orbitopathy rather than sympathetic overactivity.

Step 1 and Step 2 of the policy model's reasoning appropriately integrate the patient’s symptoms to suspect thyroid-related exophthalmos, demonstrating sound medical reasoning. However, in Step 5, the model incorrectly attributes ocular symptoms to sympathetic overactivity, ultimately leading to the selection of an incorrect final answer.

The retrieved document clarifies that diplopia and conjunctival injection are characteristic only of autoimmune orbitopathy in Graves’ disease. This distinction is key to differentiating the incorrect choice (E) from the correct one (C). Leveraging this evidence, \OURS{} assigns step-level labels of 1 to Step 1 and Step 2, and 0 to Step 5 and Step 6, aligning exactly with expert annotations. 

In contrast, the ORM data labeling pipeline assigns an overall incorrect label to the entire reasoning trace, as it only considers the final answer, missing the valid reasoning steps in Steps 1 and 2. Similarly, the conventional PRM data labeling pipeline assigns low scores to Steps 1 and 2, incorrectly marking them as poor reasoning, since the final answer derived from these steps is incorrect. As a result, both ORM and PRM fail to correctly evaluate the quality of intermediate reasoning steps, mislabeling valid reasoning as low quality.

\paragraph{Test-Time Labeling}
Figure~\ref{fig:comparison_scoring} presents an example from the MedQA dataset concerning a patient suspected of polyhydramnios. Duodenal atresia can cause polyhydramnios, while posterior urethral valve (PUV) typically leads to oligohydramnios.

Steps 1 to 4 of the reasoning trace appropriately integrate the patient’s clinical findings to analyze the polyhydramnios context. However, in Step 5, the model incorrectly reasons that posterior uretheral valve can cause polyhydramnios, ultimately leading to an incorrect answer in Step 6.

The retrieved document explicitly states that PUV leads to oligohydramnios due to urinary outflow obstruction. This serves as a clear rationale for why the incorrect answer choice (D) is invalid.

Conventional PRM models assign relatively high rewards to Step 5 and Step 6 in the absence of helpful external documents. In contrast, \OURS{} uses the retrieved document to detect the error more precisely, providing high rewards up to Step 4 and then sharply reducing rewards from Step 5 onward.

These examples show that \OURS{} can identify and localize errors in medical reasoning steps more effectively than auto-labeling trained PRM, highlighting the value of document-grounded, stepwise evaluation in capturing nuanced reasoning quality by referring to relevant clinical documents.

%% file: figs/agent_clinic.tex

%% file: figs/labeling_strategies_comparison.tex
\begin{figure*}[t]
\centering
\includegraphics[width=0.85\textwidth]{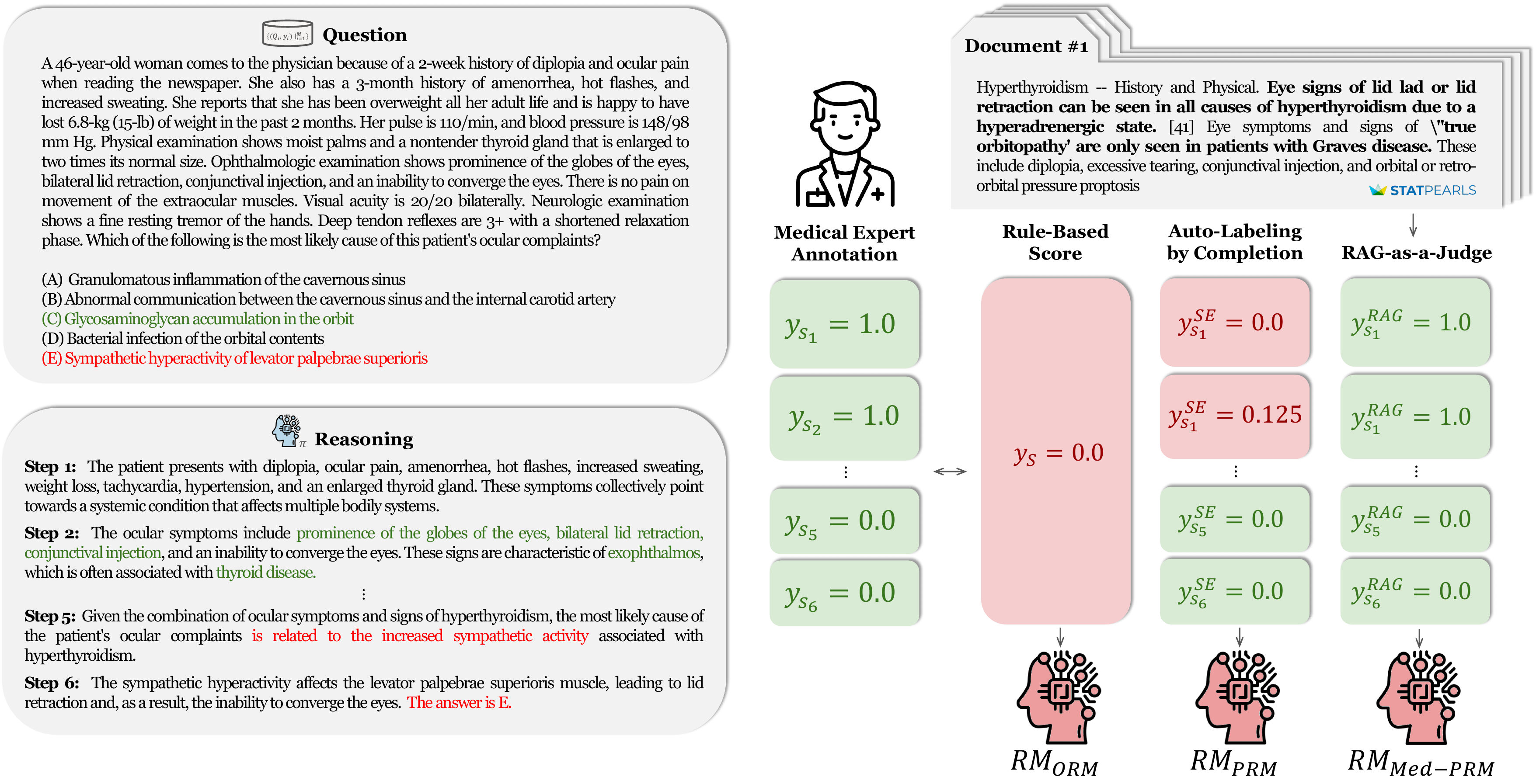}
\caption{Case study comparison of labeling strategies for reward model training.
This example illustrates how MED-PRM labeling yields more clinically accurate judgments than both rule-based and auto-completed annotations.}
\label{fig:labeling_strategies}
\end{figure*}

%% file: tables/human_annotation.tex
\begin{table}[t]
\centering
\resizebox{\columnwidth}{!}{%
\begin{tabular}{cccc}
\toprule
\textbf{Subset} & \textbf{\OURS{}} & \textbf{Soft label} & \textbf{Hard label} \\
\midrule
Easy & \textbf{0.74} & 0.64 & 0.70 \\
Hard & \textbf{0.71} & 0.34 & 0.31 \\
\bottomrule
\end{tabular}
}
 \caption{Pearson correlation between model-generated labels and human annotations on reasoning steps for easy and hard subsets.}
\label{tab:human_annotation}
\end{table}

%% file: figs/comparison_scoring.tex
\begin{figure*}[htbp]
\centering
\includegraphics[width=0.85\textwidth]{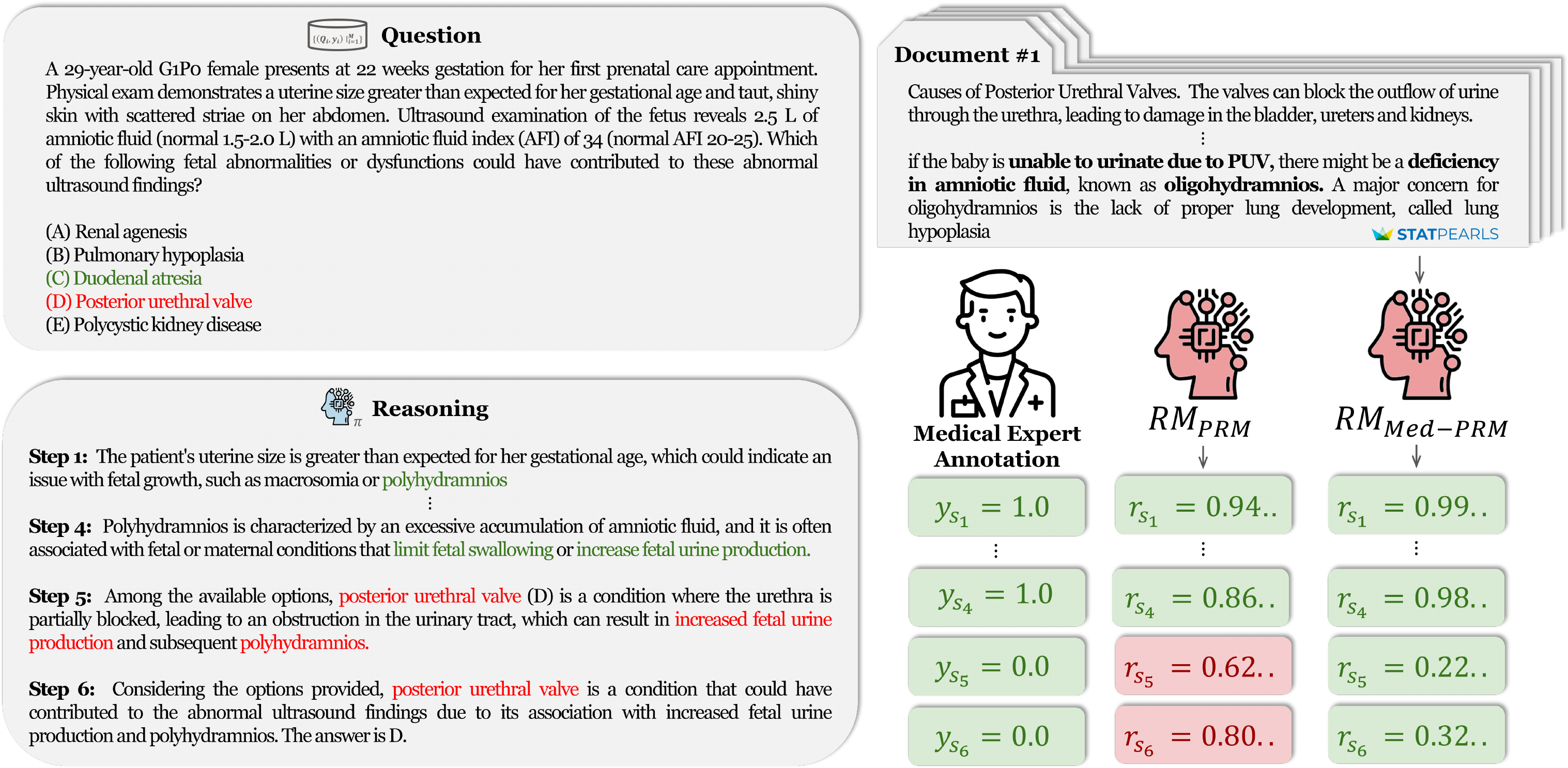}
\caption{Case study comparison of step-wise reward assignment between conventional PRM and \OURS{}. \OURS{} assigns more appropriate and clinically grounded rewards than conventional PRM by leveraging evidence from retrieved documents, enabling more accurate evaluation of intermediate reasoning steps.}
\label{fig:comparison_scoring}
\end{figure*}

%% file: tabs/08_conclusion.tex
In this study, we present \OURS{}, a process reward modeling framework that advances medical reasoning in language models with \RAGLLM{} approach. \OURS{} addresses key limitations of existing PRMs by verifying intermediate reasoning steps against retrieved medical documents rather than relying on outcome-based auto-labeling. Our experiments show that \OURS{} consistently outperforms prior methods across both multiple-choice and open-ended clinical benchmarks, with notable gains in complex diagnostic tasks. Furthermore, the retrieval-based step verification produces scalable, expert-aligned labels that support effective policy training and verifier in test-time scaling. Together, these contributions establish \OURS{} as a robust and generalizable approach for improving the accuracy, transparency, and trustworthiness of medical AI systems.

%% file: tabs/99_limitations.tex
Our work has a few limitations that warrant discussion. First, our evaluation is currently confined to the medical domain, though the methodology could potentially generalize to other domains requiring stepwise reasoning verification and retrieval-augmented generation. Second, due to computational constraints, we limited our experiments to small language models like Llama 3.1 8B and Meerkat 8B, though there remains significant potential to explore scalability across different model architectures and sizes Third, while our reward model demonstrates strong performance, we did not extensively explore diverse reinforcement learning methods that could further enhance our method's capabilities. Future work should investigate these aspects through broader domain coverage, model scaling experiments, and more sophisticated reinforcement learning training strategies.

%% file: tabs/99_acknowledgments.tex
This research was supported by the National Research Foundation of Korea (NRF-2023R1A2C3004176), the Ministry of Health \& Welfare, Republic of Korea (HR20C002103), the Ministry of Science and ICT through Seoul National University Hospital (RS-2023-00262002), and the ICT Creative Consilience Program through the Institute of Information \& Communications Technology Planning \& Evaluation (IITP) grant funded by the Korea government (MSIT) (IITP-2025-RS-2020-II201819).

%% file: tabs/08_appendix.tex
\section{Notation}
\label{appendix:notation}
\input{tables/notation}
Table~\ref{tab:notation} summarizes the notations used throughout the manuscript for the convenience of readers based on \citet{math_shepherd_wang}

\section{Retrieval}
\label{appendix:retrieval}
We use MedCPT~\cite{jin_medcpt_2023} bi-encoder and cross-encoder for dense retrieval and reranking from four biomedical corpora:
\begin{itemize}
    \item Clinical Guidelines~\cite{chen_meditron-70b_2023}
    \item StatPearls~\cite{xiong_benchmarking_2024}
    \item Medical Textbooks~\cite{singhal_towards_2023}
    \item Rare Disease Corpus~\cite{wang_assessing_2024}
\end{itemize}
Retrieval is performed on AWS EC2 c5.9xlarge with Milvus~\cite{wang_milvus_2021} using Max Inner Product Search (MIPS); reranking uses an NVIDIA RTX 3090. For each query, 100 documents per corpus (400 total) are retrieved, with top 32 selected after reranking, following~\cite{sohn_rationale-guided_2025}.

\section{Hyperparameters}
\label{appendix:hyperparameters}
\input{tables/hyperparameters}
Table~\ref{tab:hyperparameters} shows the hyperparameters used for training our models of \OURS{} framework

\section{API Usage and Cost}
We utilized the Gemini-2.0-flash model via the Google Generative AI API. The model was employed as a \RAGLLM{}, specifically for generating training labels for the PRM and for scoring responses in the AgentClinic benchmark. All API calls were made using the official endpoint, with the temperature set to 0 and standard rate limits applied. The total API cost incurred for curating the training set of PRM was approximately \$20.

\section{Related Works} \label{supp:related}
\paragraph{Medical Reasoning}
Large language models (LLMs) have shown growing competence in medical reasoning, which presents unique challenges beyond general language tasks, including specialized terminology, multimodal patient data, and high demands for factual accuracy. Leveraging biomedical corpora such as PubMed, MIMIC-III/IV~\cite{johnson_mimic-iv_2023}, and UMLS, recent models have progressed from surface-level recall to more complex diagnostic and therapeutic inference. Pioneered by Med-PaLM~\cite{singhal_large_2023},  demonstrates strong performance on expert-level medical questions.

Advances in Chain-of-Thought (CoT) prompting and training~\cite{wei_chain--thought_2023, xu_mental-llm_2024, kim_small_2025}, Retrieval-Augmented Generation~\cite{zakka2024almanac, sohn_rationale-guided_2025, zheng_miriad_2025}, agentic systems~\cite{kim_mdagents_2024, tang_medagents_2024, schmidgall_agentclinic_2024, tang_medagentsbench_2025} further extend this paradigm by orchestrating multiple specialized agents to collaboratively solve complex clinical tasks. PRM~\cite{lightman_lets_2023, jiang_meds3_2025}, and Reinforcement Learning~\cite{math_shepherd_wang} have enabled LLMs to generate structured reasoning traces. Similarly, HuatuoGPT-o1~\cite{chen_huatuogpt-o1_2024} incorporates verifier feedback to iteratively refine multi-step reasoning traces.

These trends highlight the growing importance of process reward mechanisms. Like mathematical problem solving, clinical reasoning often requires multi-step inference, where a single incorrect step can invalidate the final outcome. Thus, stepwise supervision is not only applicable to medicine but also critical for ensuring transparency and reliability in clinical decision-making.

\paragraph{PRM in Medical Domain}
A recent study has been made to apply PRM in the medical domain. MedS$^3$~\cite{jiang_meds3_2025} constructs a PRM training dataset with similar approaches to existing PRM frameworks, using MCTS-based auto-labeling. \OURS{} also provides process rewards for medical reasoning, but mainly differs in the way of constructing the training set, leveraging retrieval-augmented generation and LLM-as-a-Judge. Section~\ref{sec:results} and Section~\ref{sec:analysis} show that our approach is more effective than MCTS-based methods.

\paragraph{LLM-as-a-Judge for Reasoning Evaluation}
Recent research actively explores the use of LLM-as-a-Judge as a PRM or as a Process Explanation Model (PEM). In \citet{hao_llm_2024}, LLMs are employed to evaluate the quality of Chain-of-Thought (CoT) reasoning. \OURS{} uses \RAGLLM{} as a labeling strategy for reasoning steps, which differs in terms of how it is utilized, and this also contrasts with the automatic labeling approaches used in prior PRM research.

\paragraph{Retrieval-Augmented PRM in Mathematics}
In the mathematical domain, a recent work has explored combining retrieval with PRM~\cite{zhu_retrieval-augmented_2025}. RAG-PRM retrieves semantically similar sets of questions and answers to enable PRM to generalize against out-of-distribution questions. While both \OURS{} approaches incorporate retrieval, \OURS{} differs in that it retrieves medical evidence and knowledge rather than similar QA pairs for few-shot-style prompting. Moreover, our retrieval method supports scalable integration, ranging from curated sources such as medical textbooks and clinical guidelines to potentially broader corpora like PubMed in other deployments.

\section{Benchmarks}
\label{appendix:benchmarks}
Below are detailed descriptions of the medical benchmark datasets used in our study.

\paragraph{MedQA} MedQA is a comprehensive medical question answering dataset derived from professional medical board examinations. The dataset spans three languages: English, simplified Chinese, and traditional Chinese. Our work focuses exclusively on the English subset, which contains 12,730 questions from the United States Medical Licensing Examination (USMLE). Our method was evaluated on questions with both four and five options. MedQA evaluates diverse aspects of medical knowledge, encompassing diagnostic procedures, treatment protocols, and fundamental medical concepts. The questions are designed to test both factual medical knowledge and clinical reasoning capabilities.

\paragraph{MedMCQA} MedMCQA \cite{pal_medmcqa_2022} is an extensive multiple-choice dataset comprising over 194,000 high-quality questions from Indian medical entrance examinations (AIIMS and NEET PG). Our evaluation incorporates 500 questions from this dataset, with an average token length of 12.77 tokens per question. Each question presents four answer choices. The dataset demonstrates remarkable topical diversity, covering 2,400+ healthcare topics across 21 medical subjects. MedMCQA's comprehensive coverage and focus on entrance exam questions make it particularly valuable for assessing both theoretical knowledge and practical clinical reasoning abilities in medical problem-solving scenarios.

\paragraph{MMLU (Medical Subset)} The Massive Multitask Language Understanding benchmark \cite{hendrycks_measuring_2021} contains specialized medical knowledge subsets that we incorporate into our evaluation. Our benchmark utilizes 1,089 medical-related questions from MMLU. Each question presents four multiple-choice options. The medical subsets encompass diverse domains including anatomy, clinical knowledge, college medicine, medical genetics, and professional medicine. MMLU's comprehensive coverage spans both fundamental and advanced medical concepts, testing knowledge across a wide spectrum of difficulty levels from basic to professional expertise. The benchmark's standardized assessment format enables meaningful comparisons between medical reasoning capabilities and other knowledge domains.

\paragraph{DDXPlus} DDXPlus~\cite{tchango_ddxplus_2022} is a large-scale synthetic dataset containing approximately 1.3 million patient cases, designed to advance research in Automatic Symptom Detection (ASD) and Automatic Diagnosis (AD) systems. Unlike traditional medical datasets that only include binary symptoms and antecedents, DDXPlus incorporates categorical and multi-choice symptoms, along with hierarchical symptom organization. Each case includes comprehensive information such as differential diagnoses, ground truth pathologies, symptoms, and relevant antecedents. This dataset enables the development of more sophisticated medical reasoning systems that can interact with patients in a logical manner and provide differential diagnoses, which is crucial for helping doctors understand the reasoning process of AI systems.

\paragraph{AgentClinic*} AgentClinic~\cite{schmidgall_agentclinic_2024} is a multimodal agent benchmark designed to evaluate large language models (LLMs) in simulated clinical environments. Unlike traditional static question answering benchmarks, AgentClinic captures the complex, sequential nature of clinical decision making by integrating diverse clinical findings derived from patient interactions, multimodal data collection, and tool usage. The benchmark spans nine medical specialties and seven languages, providing a comprehensive evaluation framework. Notably, when MedQA problems are presented in AgentClinic’s sequential decision making format, diagnostic accuracies can drop significantly compared to traditional formats. The benchmark enables novel patient-centric metrics and supports various tools including experiential learning, adaptive retrieval, and reflection cycles. AgentClinic’s interactive environment allows for in-depth evaluation of clinical reasoning capabilities through real-world electronic health records and clinical reader studies.

In this work, we adopt a simplified variant of the benchmark, referred to as AgentClinic*, which reformulates the task into a single step inference problem. This adaptation is motivated by practical considerations: conducting multiple reasoning steps would incur excessive API calls and computational overhead in large-scale experiments. Moreover, techniques like self-consistency, which are important for evaluating model reliability, are less applicable in multi-turn settings due to non-deterministic agent trajectories. AgentClinic* thus strikes a balance between realism and tractability while preserving the core challenge of evidence-grounded clinical reasoning.

\paragraph{Training}
We train our model using four widely adopted medical datasets: MedQA, MedMCQA, PubMedQA, and MMLU-Med. For MedQA, we utilize the entire training set comprising 10,178 questions. For the remaining three datasets, we randomly sampled 500 examples from each training set to construct a lightweight yet diverse training corpus, encompassing a variety of question formats and clinical topics. This setup ensured data efficiency while enabling the model to learn from a broad range of medical problem types. 

\paragraph{Evaluation}
Model performance was evaluated on MedQA, MedMCQA, PubMedQA, and six medical-related subsets of MMLU (Anatomy, Clinical Knowledge, College Biology, College Medicine, Medical Genetics, and Professional Medicine). Additionally, we conducted out-of-domain evaluations that required more complex clinical reasoning. These included DDXPlus and two variants of AgentClinic based on NEJM case reports and MedQA. For DDXPlus, we randomly sampled 500 examples due to its extensive size, and reformulated the task to focus on differential diagnosis by providing supporting evidence and requiring the model to select the correct disease from a list of up to five candidates. AgentClinic, in contrast, presented an open-ended question-answering format without predefined answer choices, simulating real-world clinical scenarios through the analysis of diverse clinical findings (multi-turn dialogues were not included). To evaluate responses in open-ended settings, we adopted an LLM-as-a-judge framework (Gemini-2.0-flash), following a similar approach to HuatuoGPT-o1~\cite{chen_huatuogpt-o1_2024}. 

\section{Full Ablation Study of PRMs vs SC}
\label{appendix:ablation study} 
\input{figs/ablation_appendix}
As shown in Figure~\ref{fig:ablation_appendix},  \OURS{} outperforms MedS$^3$ when scaling test-time computation.

\section{Prompts}
\label{appendix:prompts}

\begin{tcolorbox}[
  colback=gray!5!white,
  colframe=gray!75!black,
  title=Multiple Choice Questions CoT Prompt \texttt{system\_message}
]
Solve the following question step-by-step. Do not analyze individual options in a single step. Each step of your explanation must start with 'Step {number}: ' format. You must provide the answer using the phrase 'the answer is (option alphabet)' at the end of your step.
\end{tcolorbox}
\vspace{0.5em}
\begin{tcolorbox}[
  breakable,
  colback=gray!5!white,
  colframe=gray!75!black,
  title=Open-Ended Questions CoT Prompt \texttt{system\_message}
]
Solve the following question step-by-step. Each step of your explanation must start with 'Step {number}: ' format. The final answer must output a concise and clearly defined diagnostic term.  You must provide the final answer using the phrase '\#\# Final Diagnosis: {Disease name}' at the end of your final step. Please refer to the following example. \#\# Final Diagnosis: Multiple Sclerosis
\end{tcolorbox}

\begin{tcolorbox}[
  breakable,
  colback=gray!5!white,
  colframe=gray!75!black,
  title=System Message for Open-Ended Question Evaluation Using LLM-AS-A-JUDGE \texttt{system\_message},
]
The following presents a short-answer question along with its Ground Truth and the Model's Answer. Evaluate the Model's Answer strictly based on its correctness. Your output must be either 1 or 0 only. Output 1 if the answer is correct, and 0 if it is incorrect.
\end{tcolorbox}

\begin{tcolorbox}[
  breakable,
  colback=gray!5!white,
  colframe=gray!75!black,
  title=Generating PRM Training Data Labels Using \RAGLLM{} \texttt{system\_message},
]
You are an evaluator responsible for assessing the quality of **wrong solutions** to medical questions in a stepwise manner.
Each question is accompanied by relevant documents, a question, and the correct answer, and the quality of reasoning at each step must be evaluated.
Give a score of 0 if the response lacks logical coherence or is not based on medical evidence, and 1 if this is not the case.
Please note that if the explanation does not match the provided ground truth, it must be scored as 0.
Critically assess the reasoning at each step.
At the end of your evaluation, you must include a final summary of the scores in the following format:

\#\# Step 1: 0 or 1

\#\# Step 2: 0 or 1

\#\# Step 3: 0 or 1
...
\end{tcolorbox}

\begin{tcolorbox}[
  breakable,
  colback=gray!5!white,
  colframe=gray!75!black,
  title=\OURS{} \texttt{system\_message},
]
You are an evaluator assessing the logicality and validity of the reasoning in each step of the given explanation. In order to support the evaluation, the relevant documents, the question, and the explanation are provided sequentially. If the reasoning contains errors, output - after that step. If the reasoning in a step is logical and valid, output + after that step. 
\end{tcolorbox}

\begin{tcolorbox}[
  breakable,
  colback=gray!5!white,
  colframe=gray!75!black,
  title=PRM \texttt{system\_message},
]
You are an evaluator assessing the logicality and validity of the reasoning in each step of the given explanation. In order to support the evaluation, the question and the explanation are provided. If the reasoning contains errors, output - after that step. If the reasoning in a step is logical and valid, output + after that step.
\end{tcolorbox}

\section{Human Evaluation Details}
\label{appendix:human evaluation Guidelines}
The evaluation was conducted by one physician with four years of clinical experience and two senior medical students. Each of the two medical students independently annotated all reasoning steps, and all annotations—including those with disagreements—were subsequently reviewed and adjudicated by the physician. The evaluation followed the guidelines described below.

\begin{tcolorbox}[
  breakable,
  colback=gray!5!white,
  colframe=cyan!60!blue,
  title=Human Evaluation Guidelines,
]
The following content presents a stepwise explanation of a medical problem. Provide a critical evaluation of each step based on an integrated assessment of the following criteria.

\vspace{1ex}
\textbf{Evaluation Criteria}
\vspace{0.5ex}

- Factual Accuracy:Does the step accurately reflect established medical knowledge? Are there any inconsistencies or factual inaccuracies?
\vspace{1ex}

- Problem-Solving Relevance: Does the step contribute meaningfully to solving the problem? Does it avoid diverging into irrelevant or tangential reasoning?
\vspace{1ex}

- Logical Coherence: Is the reasoning based on appropriate medical knowledge and logically consistent within the clinical context?
\vspace{2ex}

\textbf{Scoring Method}
\vspace{0.5ex}

- Assign 1 point or 0 points to each step.
\vspace{1ex}

- 1 point: Awarded when the step is generally factually accurate, contributes to solving the problem, and demonstrates coherent medical reasoning.
\vspace{1ex}

- 0 points: Assigned when the step contains significant factual errors or involves reasoning that critically undermines the problem-solving process.
\vspace{1ex}

- If a step contains a critical error, any subsequent steps that rely on or are influenced by that error should also be scored as 0.
\end{tcolorbox}

\clearpage
\onecolumn
\vspace*{\fill}
    \section{Scaling Test-Time Computation with PRMs Across Multiple Models}
    \label{appendix:Scaling Test-Time Computation with PRMs Across Multiple Models}
    \input{tables/table1_appendix}
\vspace*{\fill}
\twocolumn
\clearpage

%% file: tables/notation.tex
\begin{table}[h]
    \footnotesize
    \renewcommand{\arraystretch}{1.3}
    \centering
    \resizebox{\columnwidth}{!}{%
      \begin{tabular}{@{}cl@{}}
        \toprule
        \textbf{Symbol} & \textbf{Description} \\ \midrule
        $q$                   & medical question \\
        $D$                   & retrieved documents  \\
        $K$                   & number of reasoning steps \\
        $K_j$                 & number of reasoning steps in trace $S_j$ \\
        $i$                   & step index ($1,$ ... $,K$) \\
        $N$                   & number of reasoning traces \\
        $j$                   & trace index ($1,$ ... $,N$) \\
        $C$                   & set of reasoning traces\\
        $S$                   & reasoning trace\\
        $S_j$                 & reasoning trace $j$\\
        $s_i$                 & step $i$ of reasoning trace $S$ \\
        $s_{i,j}$             & step $i$ of reasoning trace $S_j$ \\
        $y_{S_j}$             & gold label for trace $j$ \\
        $y_{s_{i}}$           & gold label for step 
        $s_{i}$ \\
        $y^{SE}_{s_{i}}$      & soft label for step 
        $s_{i}$ \\
        $y^{HE}_{s_{i}}$      & hard label for step 
        $s_{i}$ \\        
        $y^{RAG}_{s_{i}}$     & label made by \RAGLLM{} for step 
        $s_{i}$ \\                
        $y_{s_{i,j}}$         & gold label for step $s_{i,j}$ of trace $j$ \\
        $r_{S_{j}}$           & reward (sigmoid) score for trace $j$ \\
        $r_{s_{i}}$           & reward (sigmoid) score for step $s_{i}$ \\
        $r_{s_{i,j}}$         & reward (sigmoid) score for step $s_{i,j}$ of trace $j$ \\
        $a^{*}$               & gold answer of $q$\\
        $a_{S_j}$             & decoded answer of trace $j$ \\
        $a_{RM}$              & answer selected by reward model \\
        $RM$($D, q, S_j$)     & reward score of trace $j$ \\
        $HE$                  & Hard Label \\
        $SE$                  & Soft Label \\
        $SC$                  & Self Consistency \\
        \bottomrule
      \end{tabular}}%
    \caption{Summary of notations}
    \label{tab:notation}
  \end{table}

%% file: tables/hyperparameters.tex
\begin{table}[h]
    \centering
    \footnotesize
    \begin{tabular}{lrr}
    \toprule
    \textbf{Hyperparameter} & \textbf{Reward} & \textbf{Policy} \\
    \midrule
    Learning Rate & 2e-6 & 1e-5 \\
    Learning Rate Scheduler Type & cosine & cosine \\
    Warmup Ratio & 0.05 & 0.05 \\
    Batch Size & 64 & 64 \\
    Epochs & 3 & 1 \\
    Max Token Length & 4096 & 4096 \\
    Precision & bfloat16 & bfloat16 \\
     Optimizer & AdamW & AdamW \\
    \bottomrule
    \end{tabular}
    \caption{Hyperparameters used for training \OURS{}}
\label{tab:hyperparameters}
\end{table}

%% file: figs/ablation_appendix.tex
\begin{figure}[t]
    \centering
    \includegraphics[width=\columnwidth]{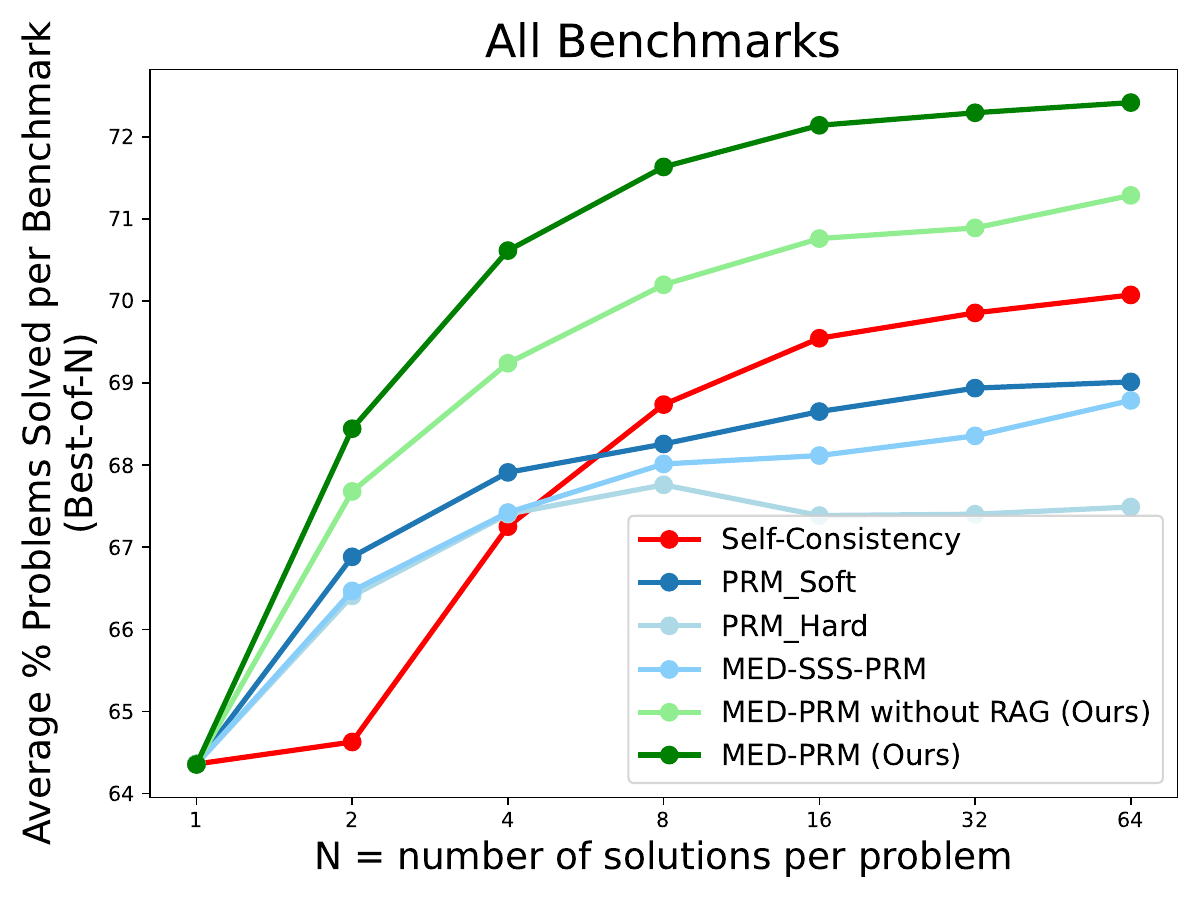}
    \label{fig:best_of_n_vs_majority}
    \caption{Comparison of scaling test-time computation performance between \OURS{} and PRMs trained with conventional approach across overall medical benchmarks(including MedS\(^3\) in the same setting).}
    \label{fig:ablation_appendix}
\end{figure}

%% file: tables/table1_appendix.tex
\begin{table}[H]
    \setlength{\tabcolsep}{1pt} 
    \scriptsize 
    \resizebox{\textwidth}{!}{
    \begin{tabular}{ll*{8}{>{\centering\arraybackslash}p{1.4cm}}}
    \toprule
    &   
    &   
    \multicolumn{5}{c}{\textbf{Multiple-Choice QA}} 
    & \multicolumn{2}{c}{\textbf{Open-Ended QA}} 
    &   
    \\  

    \cmidrule(lr){3-7} \cmidrule(lr){8-9}

    \makecell{\textbf{Category}} & \makecell{\textbf{Model}}
    & \textbf{MedQA-4} & \textbf{MedQA-5} & \textbf{MedMCQA} 
    & \textbf{MMLU-Med} & \textbf{DDXPlus} 
    & \makecell{\textbf{Agent Clinic}\\\textbf{NEJM *}} 
    & \makecell{\textbf{Agent Clinic}\\\textbf{MedQA *}} 
    & \makecell{\textbf{Average}}
    \\
    \midrule
    \multirow[c]{21}{*}{\makecell[c]{Open-source\\Language Models\\with Scaling\\Test-time\\Computation\\(Best-of-N)}}
& Llama3.1 & & & & & & & & \\
& \qquad CoT & 70.93 & 65.20 & 61.60 & 78.97 & 68.80 & 35.83 & 71.96 & 64.76 \\
& \qquad SC & \underline{74.86} & \underline{70.70} & \underline{63.40} & \textbf{81.63} & 75.20 & 49.17 & 75.23 & \underline{70.03} \\
& \qquad \textnormal{PRM\textsubscript{soft}} & 72.19 & 67.48 & \textbf{64.40} & 78.97 & 74.40 & \underline{51.67} & 73.83 & 68.99 \\
& \qquad \textnormal{PRM\textsubscript{hard}} & 71.09 & 67.48 & 61.80 & 76.49 & 70.60 & 48.33 & \underline{76.64} & 67.49 \\
& \qquad MedS$^3$ & 70.15 & 65.99 & 62.80 & 77.78 & \underline{77.60} & 49.17 & \textbf{78.04} & 68.79 \\
& \qquad \cellcolor{gray!15}\textbf{\OURS{}}
    & \cellcolor{gray!15}\textbf{76.98} & \cellcolor{gray!15}\textbf{73.06} & \cellcolor{gray!15}\underline{63.40} & \cellcolor{gray!15}\underline{81.18} & \cellcolor{gray!15}\textbf{78.00} & \cellcolor{gray!15}\textbf{57.50} & \cellcolor{gray!15}\underline{76.64} & \cellcolor{gray!15}\textbf{72.39} \\
        \cmidrule{2-10}
& UltraMedical & & & & & & & & \\
& \qquad CoT & 72.66 & 68.34 & 62.60 & 79.61 & 72.60 & 45.83 & 70.56 & 67.46 \\
& \qquad SC & 75.63 & \underline{71.80} & \underline{64.20} & \underline{81.91} & \textbf{76.20} & 49.17 & 76.64 & 70.79 \\
& \qquad \textnormal{PRM\textsubscript{soft}} & 75.65 & \underline{71.80} & 63.00 & 81.36 & 73.60 & \textbf{55.00} & \textbf{77.57} & \underline{71.14} \\
& \qquad \textnormal{PRM\textsubscript{hard}} & \underline{76.28} & 71.48 & 63.40 & 80.99 & 72.80 & 48.33 & 74.77 & 69.72 \\
& \qquad MedS$^3$ & 74.00 & 68.66 & 63.00 & 80.62 & \underline{75.80} & 48.33 & 74.77 & 69.31 \\
& \qquad \cellcolor{gray!15}\textbf{\OURS{}} & \cellcolor{gray!15}\textbf{76.42} & \cellcolor{gray!15}\textbf{74.94} & \cellcolor{gray!15}\textbf{64.40} & \cellcolor{gray!15}\textbf{83.20} & \cellcolor{gray!15}75.40 & \cellcolor{gray!15}\textbf{55.00} & \cellcolor{gray!15}\underline{77.10} & \cellcolor{gray!15}\textbf{72.35} \\
        \cmidrule{2-10}
& \OURS{} $\pi$ & & & & & & & & \\
& \qquad CoT & 67.16 & 64.26 & 57.20 & 75.48 & 67.20 & 40.00 & 68.69 & 62.86 \\
& \qquad SC & \underline{76.04} & \underline{71.80} & \underline{62.20} & \underline{82.00} & \underline{74.80} & \underline{50.83} & \underline{78.04} & \underline{70.82} \\
& \qquad \textnormal{PRM\textsubscript{soft}} & 72.58 & 68.81 & 60.20 & 78.97 & 74.60 & 49.17 & 77.10 & 68.78 \\
& \qquad \textnormal{PRM\textsubscript{hard}} & 72.27 & 69.05 & 60.40 & 78.33 & 71.40 & 45.83 & 73.83 & 67.30 \\
& \qquad MedS$^3$ & 69.60 & 67.09 & 60.00 & 78.79 & \underline{74.80} & 48.33 & 71.50 & 67.16 \\
& \qquad \cellcolor{gray!15}\textbf{\OURS{}} & \cellcolor{gray!15}\textbf{76.76} 
    & \cellcolor{gray!15}\textbf{73.06} & \cellcolor{gray!15}\textbf{64.40} & \cellcolor{gray!15}\textbf{82.46} & \cellcolor{gray!15}\textbf{77.80} & \cellcolor{gray!15}\textbf{54.17}
    & \cellcolor{gray!15}\textbf{80.37} & \cellcolor{gray!15}\textbf{72.72} \\
            \midrule
    \multirow[c]{21}{*}{\makecell[c]{Open-source\\Language Models\\with Scaling\\Test-time\\Computation\\(SC+RM)}}
& Llama3.1 & & & & & & & & \\
& \qquad CoT & 70.93 & 65.20 & 61.60 & 78.97 & 68.80 & 35.83 & 71.96 & 64.76 \\
& \qquad SC & 74.86 & 70.70 & 63.40 & 81.63 & 75.20 & 49.17 & 75.23 & 70.03 \\
& \qquad \textnormal{PRM\textsubscript{soft}} & \underline{75.49} & \underline{72.19} & 65.00 & \underline{82.28} & 75.00 & 50.83 & 76.17 & \underline{70.99} \\
& \qquad \textnormal{PRM\textsubscript{hard}} & 75.33 & 71.09 & 64.20 & 81.73 & 75.20 & 51.67 & 75.23 & 70.64 \\
& \qquad MedS$^3$ & 73.84 & 70.15 & \underline{64.40} & 81.36 & \underline{75.80} & \textbf{54.17} & \underline{76.64} & 70.91 \\
& \qquad \cellcolor{gray!15}\textbf{\OURS{}}
    & \cellcolor{gray!15}\textbf{78.24} & \cellcolor{gray!15}\textbf{73.53} & \cellcolor{gray!15}\textbf{66.40} & \cellcolor{gray!15}\textbf{83.29} & \cellcolor{gray!15}\textbf{76.80} & \cellcolor{gray!15}\underline{53.33} & \cellcolor{gray!15}\textbf{77.10} & \cellcolor{gray!15}\textbf{72.67} \\
        \cmidrule{2-10}
& UltraMedical & & & & & & & & \\
& \qquad CoT & 72.66 & 68.34 & 62.60 & 79.61 & 72.60 & 45.83 & 70.56 & 67.46 \\
& \qquad SC & 75.63 & 71.80 & 64.20 & 81.91 & 76.20 & 49.17 & 76.64 & 70.79 \\
& \qquad \textnormal{PRM\textsubscript{soft}} & \underline{77.14} & \underline{72.43} & \underline{65.40} & \underline{82.46} & 75.60 & 51.67 & \textbf{78.50} & \underline{71.89} \\
& \qquad \textnormal{PRM\textsubscript{hard}} & 76.90 & 72.19 & 64.40 & 82.37 & 75.80 & 50.00 & \underline{78.04} & 71.39 \\
& \qquad MedS$^3$ & 75.41 & 72.03 & 63.00 & 81.36 & \underline{76.60} & \textbf{54.17} & 75.23 & 71.11 \\
& \qquad \cellcolor{gray!15}\textbf{\OURS{}} & \cellcolor{gray!15}\textbf{79.87} & \cellcolor{gray!15}\textbf{75.26} & \cellcolor{gray!15}\textbf{65.50} & \cellcolor{gray!15}\textbf{82.83} & \cellcolor{gray!15}\textbf{77.60} & \cellcolor{gray!15}\underline{52.50} & \cellcolor{gray!15}77.57 & \cellcolor{gray!15}\textbf{73.02} \\
        \cmidrule{2-10}
& \OURS{} $\pi$ & & & & & & & & \\
& \qquad CoT & 67.16 & 64.26 & 57.20 & 75.48 & 67.20 & 40.00 & 68.69 & 62.86 \\
& \qquad SC & 76.04 & 71.80 & 62.20 & 82.00 & 74.80 & 50.83 & 78.04 & 70.82 \\
& \qquad \textnormal{PRM\textsubscript{soft}} & 76.51 & 72.51 & \underline{63.60} & \underline{82.55} & \underline{75.80} & \textbf{54.17} & 78.97 & \underline{72.02} \\
& \qquad \textnormal{PRM\textsubscript{hard}} & \underline{77.06} & \underline{72.58} & 63.00 & 82.19 & 75.60 & \underline{53.33} & \textbf{79.44} & 71.89 \\
& \qquad MedS$^3$ & 75.26 & 72.03 & 63.40 & 81.82 & 75.60 & 52.50 & 78.04 & 71.24 \\
& \qquad \cellcolor{gray!15}\textbf{\OURS{}} & \cellcolor{gray!15}\textbf{79.18} 
    & \cellcolor{gray!15}\textbf{75.49} & \cellcolor{gray!15}\textbf{67.40} & \cellcolor{gray!15}\textbf{83.29} & \cellcolor{gray!15}\textbf{77.20} & \cellcolor{gray!15}52.50
    & \cellcolor{gray!15}\textbf{79.44} & \cellcolor{gray!15}\textbf{73.50} \\
    \bottomrule
    \end{tabular}
    } 
    \caption{
Accuracy of open‐source language models with scaling test‐time computation. Three models were evaluated: Llama 3.1 8B Instruct, Llama‐3‐8B‐UltraMedical (best performing model below 10B parameters), and \OURS{} $\pi$, for solution sampling. The sampling outputs were assessed using Self‐Consistency (SC) and various PRM methods under both Best‐of‐N and SC+RM settings. Across different scaling test-time computation strategies for each sampling, the best scores are shown in \textbf{bold}, and second-best scores are \underline{underlined}. \OURS{} achieved the highest average score across test-time computation scaling methods.
}
    \label{tab:table_prm_all}
\end{table}